\documentclass[sigconf]{acmart}
\usepackage{enumitem}
\usepackage{lipsum}
\usepackage[symbol]{footmisc}

\setlist[itemize]{leftmargin=*}
\setlist[enumerate]{leftmargin=*}

\def\BibTeX{{\rm B\kern-.05em{\sc i\kern-.025em b}\kern-.08emT\kern-.1667em\lower.7ex\hbox{E}\kern-.125emX}}
    
\copyrightyear{2019}
\acmYear{2019}
\acmConference[RecSys 19]
{CARS Workshop}
{CARS Workshop}
{Copenhagen, Denmark}

\begin{document}

%
\title{Deep Context-Aware Recommender System Utilizing Sequential Latent Context}

%

\author{Amit Livne}
\authornotemark[7]

\email{livneam@post.bgu.ac.il}
\affiliation{%
  \institution{Ben-Gurion University of the Negev and Telekom Innovation Laboratories at BGU}
  \streetaddress{P.O. Box 84105}
  \city{Beer-Sheva}
  \state{Israel}
  \postcode{43017-6221}
}

\author{Moshe Unger}
\authornotemark[7]
\email{munger@stern.nyu.edu}
\affiliation{%
  \institution{Stern School of Business, New York University}
  \streetaddress{P.O. Box 84105}
  \city{New York}
  \state{USA}
}

\author{Bracha Shapira}
\email{bshapira@bgu.ac.il}
\affiliation{%
  \institution{Ben-Gurion University of the Negev and Telekom Innovation Laboratories at BGU}
  \streetaddress{P.O. Box 84105}
  \city{Beer-Sheva}
  \state{Israel}
}

\author{Lior Rokach}
\email{liorrk@bgu.ac.il}
\affiliation{%
  \institution{Ben-Gurion University of the Negev and Telekom Innovation Laboratories at BGU}
  \streetaddress{P.O. Box 84105}
  \city{Beer-Sheva}
  \state{Israel}
}

\renewcommand{\shortauthors}{Livne and Unger, et al.}

%

\begin{abstract}
\footnotetext[7]{Both authors contributed equally to the paper}
Context-aware recommender systems (CARSs) apply sensing and analysis of user context in order to provide personalized services. Adding context to a recommendation model is challenging, since the addition of context may increases both the dimensionality and sparsity of the model. Recent research has shown that modeling contextual information as a latent vector may address the sparsity and dimensionality challenges. 
We suggest a new latent modeling of sequential context by generating sequences of contextual information and reducing their contextual space to a compressed latent space. We train a long short-term memory (LSTM) encoder-decoder network on sequences of contextual information and extract \textit{sequential latent context} from the hidden layer of the network in order to represent a compressed representation of sequential data. We propose new context-aware recommendation models that extend the neural collaborative filtering approach and learn nonlinear interactions between latent features of users, items, and contexts which take into account the sequential latent context representation as part of the recommendation process. We deployed our approach using two context-aware datasets with different context dimensions. Empirical analysis of our results validates that our proposed sequential latent context-aware model (SLCM), surpasses state of the art CARS models. 
\end{abstract}

%
%



%

%

%
\maketitle

\section{Introduction}
Context-aware computing was first defined by \cite{billsus2002adaptive} as "software that adapts according to its location of use, the collection of nearby people and objects, as well as changes to those objects over time." As such, context-aware services should react to the users' given contexts and adapt their services accordingly. The emergence and penetration of smart mobile devices have given rise to the development of context-aware systems that utilize sensors to collect data about users in order to improve and personalize services \cite{perera2014context}. 

In context-aware recommender systems (CARSs), contextual factors are taken into account when modeling user profiles and generating recommendations. 
Many studies incorporate explicit contextual information \cite{adomavicius2015context}, referred to as conditional factors, about the ratings given. The specific contexts describe the circumstances of the information collection, e.g., weather conditions ("sunny," "cloudy," 
etc.) or precise location conditions ("at home," "at work," etc.). While in most studies the set of contexts is both small enough to handle and sufficient to prevent sparsity, such context sets do not necessarily represent an optimal set of features for the recommendation process. Recent research has shown that 
latent contextual information 
can improve recommendation accuracy. In \cite{unger2016towards} the authors represent environmental features as low-dimensional unsupervised latent contexts that were extracted by an auto-encoder (AE). While this representation takes the current context of the user into account and can greatly improve recommendation accuracy, it does not model evolution of context over time. Recent studies \cite{quadrana2018sequence, chen2018sequential, tan2016improved, devooght2017long}
have shown that incorporating sequential information that models the users' behavior over time improves the quality of recommendations. 

For session-based recommender systems \cite{quadrana2018sequence}, sequences are commonly used to model the behavior of the users over time. In this setting, the recommender system (RS) makes recommendations based mainly on the behavior of the user in the current browsing session and uses the sequence of the user's actions to predict the next step and provide recommendations accordingly. 
While most studies considered the sequence of items that each user interacted with or the user's sequence of actions (such as click, like, buy, etc.) to recommend the next item \cite{soh2017deep, tan2016improved,devooght2017long}, we suggest modeling the context sequences that led to the users' current context and preferences, in order to enhance the recommendation model. We maintain user privacy by ignoring the user ID and creating context sequences observed in the system. Hence, context sequences do not model context patterns of a specific individual user and contain context information related to multiple users.

Matrix factorization (MF), which projects users and items into a shared latent space, is a common approach for latent factor model-based recommendation. Most research efforts in CARSs have been devoted to the enhancement of MF \cite{baltrunas2011matrix,unger2016towards}. Despite the effectiveness of MF for collaborative filtering, the main limitations of MF are its use of a fixed interaction function of an inner product to estimate the complex structure of user-item interactions and its use of squared regression loss for optimization, which may be suboptimal for item recommendation with implicit feedback \cite{rendle2009bpr}. We aim to address these limitations, and in this research we extend the neural network based collaborative framework (NCF) \cite{he2017neural} which suggests utilizing deep neural networks (DNNs) to learn a nonlinear function of user-item interactions. NCF is a generalized framework of classic MF, which learns the interaction function between users and items using DNNs, as opposed to the  simple and fixed linear inner product used in MF. In our approach, we suggest enhancing the recommendation with several representations of context in order to capture complex interactions between user, item, and context in both explicit and implicit manners. 

We focus on the representation of rich contextual information and suggest modeling the preferences of users for CARSs as a sequence of recent contexts. We believe that contextual information usually refers to dynamic attributes which may change when a specific activity (e.g., visiting a friend, watching a movie, etc.) is performed repeatedly \cite{zheng2015revisit}, such as the scenarios of the activities (such as time and companion) and dynamic factors from users (such as emotional states). The idea behind modeling contextual information as a sequence of recent contexts is that users' preferences can be better explained by understanding the contexts that led to the current behavior and preferences. 
We suggest utilizing a specific type of RNN, an encoder-decoder long short-term memory (LSTM) network \cite{hochreiter1997long}, in order to extract a compressed representation of the context space in an unsupervised manner called \textit{"sequential latent context"}. We performed sensitivity analysis to 
examine the effect of the sequence length on the recommendation accuracy.  

We investigate the effect of utilizing sequential latent contexts in the recommendation process and propose several neural context-aware recommendation models based on explicit and latent representations of context data derived from various context dimensions (e.g., time, location,
mood, etc.). We compare our sequential latent context-aware model (SLCM) to several context-aware algorithms using two real-world
context-aware datasets from two domains: movies and points of interest (POIs). Specifically, we compare our models' performance to extensions of NCF with explicit and latent contexts inspired by context modeling used in state of the art context-aware algorithms \cite{unger2016towards,baltrunas2011matrix}. We evaluate the accuracy of each of the recommendation models using the following metrics \cite{Said:2014:CRS:2645710.2645746}: hit@k, RMSE (root mean square error), and MAE (mean absolute error). The experiments show that our suggested models outperform traditional CARSs in all measures.

The following are the major contributions of the paper. First, we present a novel contextual modeling approach that uses latent representations of sequences of contextual information. This representation is extracted from LSTM encoder-decoder networks and is capable of modeling the evolution of contextual information over time, and more accurately captures the user's current preferences than state of the art CARSs. Second, we present several new NCF extensions for CARSs domain that improve traditional and neural MF, and state of the art context-aware recommendation methods. Our suggested models utilize contextual information in an explicit or implicit manner and learn interaction functions between users, items, and contexts. The first is an explicit neural context-aware model (ENCM) which incorporates pre-defined explicit contexts in the neural model. The second is a latent neural context-aware model (LNCM) which uses a latent representation of short-term contextual information derived by an AE network. The third is a sequential latent neural context-aware model (SLCM) which utilizes sequences of contextual information that are extracted from an LSTM encoder-decoder network. We compare and identify the usage scenarios for which each variation is preferred. 


The rest of this paper is structured as follows: section \ref{sec:related} describes related work, and section \ref{sec:method} presents our neural context-aware recommendation models that suggest integrating several representations of context into CARSs. These representations are explicit or latent contexts derived from encoder-decoder networks such as AE and LSTM encoder-decoder networks. In section \ref{seq:experiments} we describe our field experiments and provide an evaluation of the results. Finally, in section \ref{sec:conclusion} we conclude and outline plans for future research.

\section{RELATED WORK}
\label{sec:related}

The CARS domain deals with modeling and predicting users' tastes and preferences by incorporating available contextual information into the recommendation process. Prior research has shown that adding explicit or latent context to the recommendation process can improve recommendation accuracy \cite{unger2016towards,baltrunas2011matrix}. 
While the latent representation can greatly improve recommendation accuracy, it only considers a single context instance of the user. 

Despite the effectiveness of MF \cite{koren2008factorization} as the primary latent factor model of introducing users' preferences into recommendation systems, it has some limitations:
1) the multiplication of latent features linearly may not be sufficient to capture the complex structure of user interaction data, and 2) the dimensionality of the latent factor space must be explicitly defined. In order to address these limitations, \cite{he2017neural} suggested a deep neural network architecture for learning the interaction function between users and items.
We design novel context-aware models which include the context dimension in order to the learn nonlinear interactions between latent representations of user, item and context. We suggest several explicit and latent context representations for this task. The latent representations are obtained by utilizing encoder-decoder neural networks (AE and LSTM encoder-decoder networks) which reduce the high-dimensional context space and reveal complex correlations within the data.   

\cite{smirnova2017contextual} proposed a context-aware session-based RS utilizing conditional RNNs which injects contextual information into input and output layers and modifies the behavior of the RNN by combining context embedding with item embedding. While their work strengthens the hypothesis that contextual sequential modeling improves recommendation accuracy, they only integrated three types of explicit contextual features: time, time difference since last event, and event type. In our work, we define context as any user situation (physical and emotional) that might affect the user's preference. We use many context dimensions (e.g., weather, sound, light, location) in order to extract latent contexts that are not predefined to a limited list of activities. We assume that there are many different types of contexts, thus the task of modeling latent context is more complicated.

Denoising AEs were used by \cite{wu2016collaborative} to perform top-k recommendation by modeling distributed representations of the users and items. We use LSTM encoder-decoders in order to model latent representations of contextual information. We learn a function of user-item interaction that is related to this context information. A deep sequential recommendation model which learns long-term dependencies between user-item interactions without considering context dimensions was proposed by \cite{soh2017deep}. 
We suggest using encoder-decoder networks to extract latent representations of context based on historical dependencies of context interactions, focusing on contextual modeling which is a far more complex problem. 

Sequence-to-sequence (seq2seq) learning, proposed by \cite{kalchbrenner2013recurrent,sutskever2014sequence,cho2014learning}, is an effective paradigm for dealing with variable length inputs and outputs. Seq2seq learning uses recurrent neural networks (i.e., LSTM encoder-decoder networks) to map variable length input sequences to variable length output sequences. 
Despite the advantages of seq2seq learning for time series data, and its ability to compress variable length sequences into fixed length representations, it has not yet been used for modeling context in RSs. In our research, we apply the seq2seq learning technique on LSTM encoder-decoders and AEs (unsupervised learning). We stacked one-to-many architecture (decoder) on top of the many-to-one (encoder) (see figure \ref{fig:LSTM}). This architecture mimics many-to-many behavior which allows compressing sequences of high-dimensionality contextual information into a singular and low-dimensionality representation. By extracting latent context from the compressed layer in LSTM encoder-decoder networks or AEs, we learn implicit relations between the contextual features, which are nonlinear to the number of features and allows diversity in the learned patterns.

In our research, we extend the deep recommendation model presented in \cite{he2017neural}, proposing several context-aware approaches that use various representations of context and showing that modeling context in an implicit manner as a sequence can greatly improve the recommendation accuracy. In our extension of the model, we use the encoding-decoding process to learn latent context representations, utilize sequences of user data derived from contextual conditions to improve the accuracy of CARSs, and automatically select a small set of the best features to handle in order to address the sparsity challenge. A detailed description of our extended model is provided in section \ref{sec:method}.

\section{METHOD}
\label{sec:method}

\begin{figure}
\centering
\includegraphics
[height=2.5in, width=3.6in]{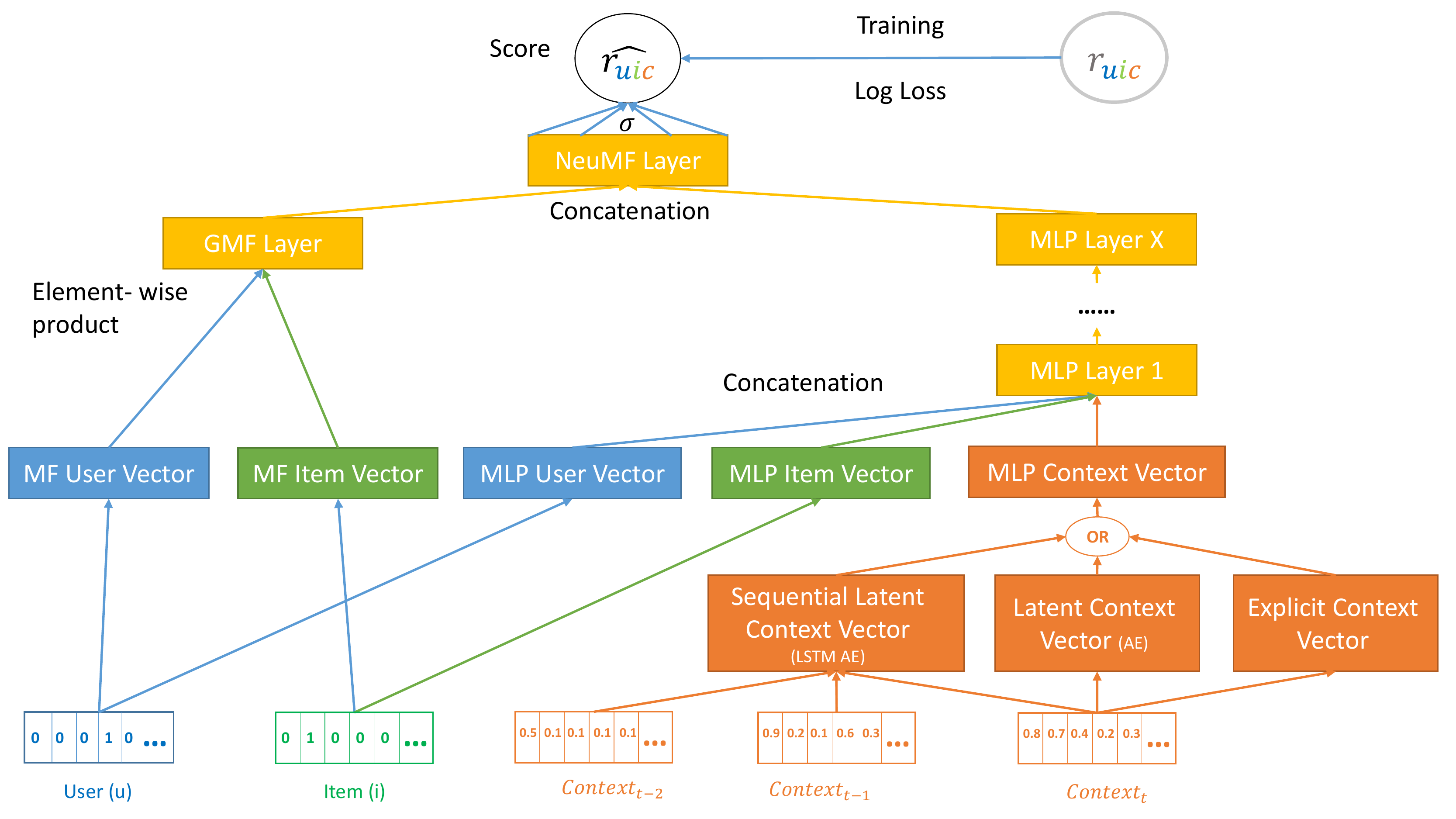}
\caption{Method Overview.}
\label{fig:method}
\end{figure}
We aim to design a novel context-aware deep recommendation model that automatically learns the relations between users, items, and contextual information. Specifically, we present novel NCF extensions models for the CARS domain that improve traditional MF. These models utilize contextual information in an explicit and latent manner and learn interaction functions between users, items, and contexts. Our method is composed of two main tasks: context extraction and its utilization in deep context-aware recommendation , as illustrated in figure \ref{fig:method}. A description of each of the tasks is provided in the following sub-sections.

\subsection{Context Extraction}
\label{sec:contextExtraction}

The context extraction
task is responsible for extracting a reduced representation of context from several contextual dimensions (i.e., mobile sensors, weather, 
etc.). We differentiate between explicit and latent context.
\begin{itemize}
    \item Explicit context describes known user situations from a predefined set of context conditions (e.g., "at work," "running," etc.) and hence can be better explained. However, it is challenging and a resource-demanding task to define and train a large enough set of explicit contexts to cover the potentially large variety of user behaviors. 
    \item Latent contexts are comprised of an unlimited number of hidden context patterns which are harder to explain. On the other hand, these patterns are modeled as numeric vectors and can be obtained automatically by applying unsupervised learning techniques.
\end{itemize}

\subsubsection{Explicit Context Extraction}
The extraction of explicit context is usually preformed by domain experts who select the most suitable context conditions for the recommendation process. The explicit context extracted reflects a single context of the user, during which he/she interacted with an item. However, context dimensions are predefined in a process that cannot be automated. 
We selected the following explicit context conditions which are common explicit factors used in CARSs \cite{baltrunas2011matrix}: "time," "location," and "day of week". 

\subsubsection{Latent Context Extraction}
\label{sec:latentContextExtraction}
The latent context extraction used in this study was derived by two encoder-decoder models, the 
AE and LSTM encoder-decoder models, as shown in figures  \ref{fig:LSTM}. These models can significantly reduce the dimensionality and noise of the feature space, and reveal relationships between the contextual features. By comparing these two unsupervised encoder-decoder models, we can determine whether users' preferences can be better understood by looking at a broader view of their context situations rather than just considering a single context instance. Unlike AEs, LSTM encoder-decoders can use their internal memory to process arbitrary sequences of inputs. Thus, we suggest identifying the dynamic temporal context 
and considering the time factor and previous implicit contexts 
in a recommendation system as follows:

  \begin{itemize}
  \item \textbf{Current context}: this latent representation is generated from the compressed layer of a trained AE 
  . It is an automatic extraction process which reflects the current situation of the user, when he/she rates an item in a RS. The auto-encoding process is performed by an unsupervised learning algorithm that applies backpropagation which is aimed at setting the value of the targets so they are equal to the value of the inputs \cite{druzhkov2016survey}. In the training phase, the encoder tries to learn the function
$h_{W,b}(\overrightarrow{Context_t}) = {\overrightarrow{\hat {Context_t}}} \simeq \overrightarrow{Context_t}$, where W and b are the weights of the network's edges. Since the current context of a user ($\overrightarrow{Context_t}$) often consists of high-dimensional feature space (such as location, 
accelerometer data, etc.) that is relatively correlated, the compressed representation can discover nonlinear correlations between the different features.
  
  \item \textbf{Sequential context}: this latent representation
  models the evolution of "current context" over time by learning long-term context patterns from sequences of context. Unlike current context which is generated by AEs, LSTM encoder-decoders can use their internal memory to process arbitrary sequences of inputs. We extract a sequential latent context from each sequence using the hidden layers of the LSTM encoder-decoder network. This process reduces the dimensionality of a sequence to a latent vector that represents a compressed representation, which is referred to as sequential latent context ($\overrightarrow{SLC}$ in figure \ref{fig:LSTM}).

 \begin{figure}[b]
\centering

\includegraphics
[width=\columnwidth]{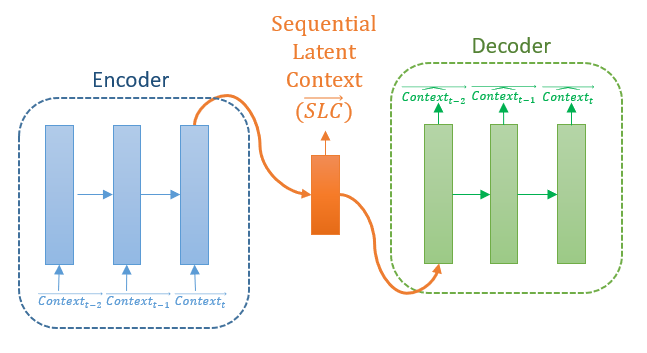}
\caption{LSTM Encoder-Decoder Model for Extracting Sequential Latent Context.}
\label{fig:LSTM}
\end{figure}
  
Our implementation of the LSTM encoder-decoder (presented in figure \ref{fig:LSTM}) is closely inspired by the one discussed in \cite{graves2013generating}. This model consists of two recurrent neural networks: the encoder LSTM and the decoder LSTM. The input to the model is a sequence of context vectors. The encoder LSTM reads this sequence, and after the last input has been read, the cell state and output state of the encoder are compressed through the hidden layers to the decoder LSTM which outputs a prediction for the target sequence. In order to compress the sequence representation, the target sequence used is the same as the input sequence. The major advantage of encoder-decoder LSTM networks is their ability to implicitly learn certain underlying characteristics of the input data. By putting constraints on the LSTM encoder-decoder, we can discover an interesting data structure, which is composed of a limited number of hidden units in each layer that force the network to learn a compressed representation of the context input. 

\end{itemize}

\subsubsection{Generating Sequences for Extracting Sequential Context}
A sequence is defined as a set of context vectors that occurred one after the other. For instance, a sequence of length three represents three interactions in the system (three vectors: $\overrightarrow{Context_{t-2}}$, $\overrightarrow{Context_{t-1}}$, and $\overrightarrow{Context_t}$ of current context that occurred at times $t-2$, $t-1$, and $t$ respectively). These interactions are collected implicitly when users interact with the system (e.g., rate an item or periodically from the mobile device). A sequence may contain context vectors from different users, since it models typical context patterns observed in the system and not context patterns of each individual user.

Both the CARS and Yelp datasets contains a time-oriented feature for each interaction, and therefore we generated sequences of context vectors by sorting them based on the timestmap feature. 
Table \ref{tab:sequncesCARS} provides a sorted example (by timestamp) of user interactions in the CARS dataset. In this example, the interactions are sorted by the time they occurred (i.e., by the timestamp feature) and represent a set of users' ratings ("like,"  "dislike," or "check-in") with contextual conditions (e.g., "light," "battery," etc.).
In order to generate sequences that are three interactions in length out of 4 sorted interactions, we generated two sequences: \{1,2,3\} and \{2,3,4\} 
(each number represents an index of the instance from table \ref{tab:sequncesCARS}).

\begin{table}[t]
\centering
\caption{An Example of Sorted Interactions for Sequence Generation with the CARS Dataset.}
\resizebox{0.5 \textwidth}{!}{
\begin{tabular}{|c|c|c|c|c|c|c|c|}
\hline
\textbf{Timestamp} & \textbf{Index} & \textbf{UserID} & \textbf{ItemID} &  \textbf{Light} & \textbf{Battery} & \textbf{...} & \textbf{Rating} \\ \hline
March 8, 2015 6:24:49 PM & 1 & 1 & 1  & 0.5 & 0.68 & ... & like \\ \hline
March 8, 2015 10:42:51 PM & 2 & 11 & 1 & 0.69 & 0.74 & ... & dislike \\ \hline
March 8, 2015 10:58:02 PM & 3 & 2 & 56 & 0.67 & 0.11 & ... & check-in \\ \hline
March 8, 2015 11:23:13 PM & 4 & 4 & 8 & 0.77 & 0.16 & ... & like \\ \hline
\end{tabular}
}
\label{tab:sequncesCARS}
\end{table}

\subsection{Deep Context-Aware Recommendation}
The second task refers to the process of training an offline recommendation model with the information collected about users, items and the extracted context for each interaction, as presented in figure \ref{fig:method}. We extend the neural matrix factorization (NeuMF) presented by \cite{he2017neural}, which we consider as our baseline model. 


In our extension of the baseline model, we propose adding a new component: a context vector (denoted by "MLP Context Vector" in figure \ref{fig:method}). This context vector is extracted efficiently from a set of contextual conditions, which can be modeled in an explicit or latent manner, as explained in section \ref{sec:contextExtraction}. Although each type of context vector is extracted in a different way, all of them share the same context vector length, which enables us to easily modify the original framework. We concatenate the MLP context vector to the MLP user vector and MLP item vector embeddings in order to learn a new nonlinear function between all three components (users, items and contexts). In this way, the context dimension is considered within the neural framework in order to automatically learn its influence on the predicted rating score $r_{u,i,c}$.

\section{EXPERIMENTS}
\label{seq:experiments}
This section is organized as follows. First, we present the datasets used in our experiments. Then, we describe the evaluation task, success metrics, baselines, and configurations of the suggested contextual models. Finally, we report and discuss the results.

\subsection{Datasets}
We implemented our models using two context-aware datasets, as presented in table \ref{tab:datasetStats}. It is important to mention that the number of available datasets with rich context data is scarce, and therefore we focused on these two datasets that contain rich contextual information and up to 4\% missing context values.   
\begin{itemize}
\item \textbf{CARS} is a high-dimensional context-aware dataset \cite{unger2018inferring} that contains three types of user ratings (i.e., like, dislike, and check-in) regarding POIs. For each user interaction, the dataset contains 247 contextual conditions derived by various types of sensors from the user's mobile phone, such as environmental information, user activity, mobile state, and user behavioral data. Overall, there are 38,900 
ratings (11,484 likes, 27,058 dislikes, and 358 check-ins).


\item \textbf{Yelp} \footnote{https://www.yelp.com/dataset/challenge} is a publicly available context-aware dataset that contains 5 context dimensions. Since the original data was highly sparse, we retained users and items with at least 10 interactions. This results in a subset of data that contains 150,770 users, 60,852 items, and 588,253 interactions. We use the following contextual factors: year, month, day of the week, and city. The time-based contextual factors were extracted from the date feature. We extracted further context features such as isHoliday and isWeekend from the date and location features.  
\end{itemize}

\begin{table}
\caption{Description of Context-Aware Datasets.}
\resizebox{0.5\textwidth}{!}{
\begin{tabular}{lcc}
\toprule
& \textbf{CARS Dataset}& \textbf{Yelp Dataset}
\\
\midrule
\textbf{\# of users} & 98 & 150,770  \\
\textbf{\# of items} & 1,918 & 60,852 \\
\textbf{\# of ratings} & 38,900 & 588,253\\
\textbf{rating scale} & "dislike (1), "like" (3), and "check-in"(5) & 1-5\\
\textbf{rating sparsity} & 96.41 & 99.99 \\
\textbf{\# of context dimensions} & 15 & 5 \\
\textbf{\# of context conditions} & 247 & 9\\
\toprule
\textbf{context dimensions} & time, location, ringer mode, screen, application traffic, & time, location, day type \\  & battery, activity recognition, microphone, cell state, gravity, & holiday, season \\ & light, accelerometer, orientation, magnetic field, weather  \\ \bottomrule
\end{tabular}
}
\label{tab:datasetStats}
\end{table}

We divided the data based on the record's time, using 70\% for training and the remaining 30\% for testing. The time-based division represents real-life scenarios, whereas prediction models are trained based on historical data in order to predict future samples. For both the CARS and Yelp datasets, a time-based division considered the data collected in later weeks as the test set, while the earlier weeks in the collection period were considered the training set. 

\subsection{Setup}
We evaluated the accuracy of each of the recommendation models by assessing two measures of prediction accuracy. We used RMSE which measures the difference between the actual and predicted rating, while penalizing large errors, and MAE which does not penalize large errors. For the CARS dataset we also measured the quality of the ranking recommendations and calculated the hit@k measure to evaluate the accuracy of the top-k recommendation list. A hit is defined as a POI that had a positive indication (i.e., like) by the user and was ranked among the top-k recommendations. For example, the hit@3 is one if the item that the user liked is among the top three recommended items. The list of recommendations is generated by predicting the ratings of the POIs around the user's current location (within a range of 500 meters), as inferred by the GPS, and ranking the items by their predicted rating value.

\subsubsection{Hyper-parameters}
\label{sec:hyper}
In order to train the encoder-decoder networks, we transformed nominal values of features into binary values. This process resulted in 268 features in the CARS dataset and 9 features in the Yelp dataset; for example, nominal values of type of day in the week (weekend and weekday) were transformed into two binary features (isWeekday and isWeekend). 
For each dataset we trained separate AE and LSTM encoder-decoder networks, containing three layers (input, compressed and output).
The number of units in the compressed layer, which represents the latent vector size, was determined by cross-validation in a separate calibration process for each dataset. 
For the CARS dataset, we trained the networks with 268, 40, 268 units in each layer respectively (e.g., 40 units in the compressed layer).
For the Yelp dataset, we trained the same type of networks, with 9, 10, 9 units in each layer respectively. We report the best results for the CARS dataset 
and the Yelp dataset which achieved with a sequence length of three 
and five respectively. 
For the loss function of the encoder-decoder networks, 
we used mean squared error (MSE). In order to optimize our selected loss function, we use the Adam algorithm with a squared root decay of learning rate ranging from 0.01 to 0.001. The batch size was set to 512, and the number of training iterations was set to 500.
%
%

\subsection{Baselines}
In order to test the proposed methods, we conducted a series of offline simulations to compare them to the following recommendation models:

\begin{itemize}

\item Traditional Matrix Factorization (\textit{\textbf{MF}}): we selected BiasSGD \cite{hazan2011beating}, as implemented in the GraphLab \cite{low2014graphlab} framework, which is capable of learning user/item factorization with baseline estimators.

\item Explicit Context Model (\textit{\textbf{ECM}}): contextual modeling approach suggested by \cite{baltrunas2011matrix}, which extends traditional MF and learns the rating bias under different explicit context conditions for each item. For explicit context dimensions in CARS, we chose the time and weather. For explicit context dimensions in 
Yelp, we chose season, day type, and time. 

\item Latent Context Model (\textit{\textbf{LCM}}): the latent model approach suggested by \cite{unger2016towards}, which extends traditional MF and learns the rating bias under different latent context conditions for each item. The latent contexts were extracted by an AE model, as described in section \ref{sec:latentContextExtraction}.

\item Neural Matrix Factorization (\textit{\textbf{NeuMF}}) \cite{he2017neural}: a generalization of MF to a nonlinear setting, which suggests the use of a DNNs for learning latent features of users and items, without taking into account context information. 

\end{itemize}

\subsection{Deep Context-Aware Recommendation Models}
Based on our proposed method (described in section \ref{sec:method}), we present several new NeuMF extensions for the CARS domain that include various kinds of context modeling and improve traditional MF. Specifically, we experiment with the following models:
\begin{itemize}
\item Explicit Neural Context Model (\textit{\textbf{ENCM}}): our proposed context-aware model which considers explicit context conditions. This model adds a new component of context to the NeuMF model and is modeled by a numeric vector of specific context conditions.

\item Latent Neural Context Model (\textit{\textbf{LNCM}}): our proposed context-aware model which considers low-dimensional latent context conditions. The latent representation of context conditions was extracted from the compressed layer of an auto-encoder network and added to the NeuMF model. 

\item Sequential Latent Context Model (\textit{\textbf{SLCM}}): our proposed context-aware model which considers sequential latent contexts extracted from an encoder-decoder LSTM network. The sequential latent context was taken from the compressed layer of an encoder-decoder LSTM network and added to the NeuMF model.  
\end{itemize}

\subsection{Results}

Table \ref{tab:evalResuals Hits} summarizes the performance of our suggested neural context-aware models (ENCM, LNCM, and SLCM) versus baselines in each of the datasets. 
The best results in each column are denoted in bold, while results that are statistically significant are denoted by an asterisk $(*)$.
It can be seen that in each of the datasets, our neural context-aware models outperformed the various baseline models on all of the measures. These results indicate that adding context information to the neural model improves recommendation accuracy. More specifically, in the CARS dataset SLCM achieved the best results and performed 5.6\% better than the best baseline model (NeuMF) in terms of the RMSE and 15.2\% better than the best baseline model (NeuMF) in terms of the MAE. In the 
Yelp datasets, a similar phenomenon was observed, as SLCM performed the best in terms of both the RMSE (
1.239) and MAE (
1.011) with up to 
12.3\% improvement over the best baseline (NeuMF)
.
We can also see that SLCM obtained the best results among the other suggested neural context models (i.e., ENCM and LNCM). This may be due to the fact that SLCM handled long-term context patterns which were modeled by context sequences, while ENCM and LNCM utilized only current (short-term) context of the user. 

\begin{table}
\centering
\caption{Prediction Results.}
 \resizebox{0.5 \textwidth}{!}{

\begin{tabular}{c|cc|ccccl} 
\toprule
 & \multicolumn{2}{c|}{\textbf{Yelp Dataset} } & \multicolumn{5}{c}{\textbf{CARS Dataset} } \\ 
\toprule
Model & RMSE & MAE & RMSE & MAE & Hit@1 & Hit@3 & Hit@5 \\ 
\midrule
 $MF$  & 1.921 & 1.461 & 0.487 & 0.397 & 0.117 & 0.364 & 0.507 \\
$ECM$  & 1.812 & 1.338 & 0.481 & 0.319 & 0.163 & 0.381 & 0.522 \\
$LCM$  & 1.707 & 1.287 & 0.395 & 0.301 & 0.17 & 0.374 & 0.517 \\
$NeuMF$  & 1.284 & 1.135 & 0.356 & 0.28 & 0.343 & 0.588 & 0.690 \\
$ENCM$  & 1.281 & 1.042 & 0.354 & 0.29 & 0.355 & 0.591 & 0.694 \\
$LNCM$  & 1.254 & 1.021 & 0.343 & 0.278 & 0.35 & 0.592 & 0.694 \\
$SLCM$  & \textbf{1.239*}  & \textbf{1.011}  & \textbf{0.337} & \textbf{0.243*}  & \textbf{0.399*} & \textbf{0.652*}  & \textbf{0.743*} \\
\bottomrule
\end{tabular}
}
\label{tab:evalResuals Hits}
\end{table}
Since in the CARS dataset the recommendations were presented to users as a top-10 list, we also measured the quality of ranking and performed an evaluation using various hit levels (one, three and five). It can be seen that for every value of k, the SLCM model outperforms the other models. As RSs recommend only a few items at a time, the relevant item should be among the first few items on the list. At smaller k values ($k\leq3$), the relative improvement of the SLCM model over the baseline models is 16.3\%, 10.88\%, and 6.75\% for k values of one, three, and five respectively. As expected, for larger values of k, all of the models converge to a similar behavior and achieve a high hit rate. These trends indicate that when generating topf-k recommendations the main benefit of the sequential latent context model is obtained when k is set at a small size (e.g., one or three). 
The results confirm that neural context-aware models can be effectively exploited to generate better recommendations that are related to user's context.



\begin{figure}[t]
\centering
\includegraphics[trim=2cm 0cm 0cm 1.5cm, width=70mm,height=42mm,scale=0.75]
{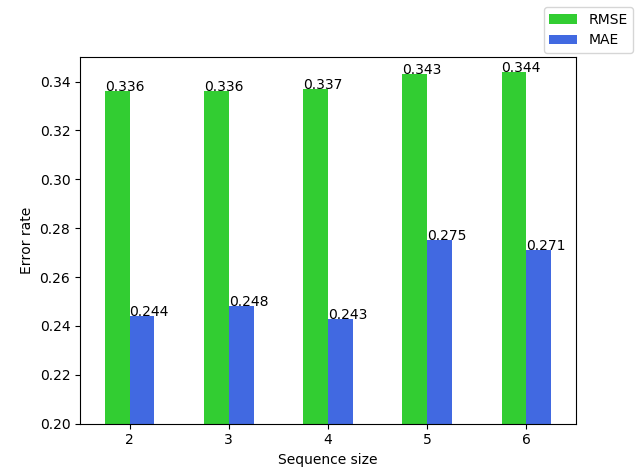}
\caption{Prediction Results of SLCM with Contextual Sequences of Different Lengths with the CARS Dataset.}
\label{fig:RMSE_SLCM}
\end{figure}

Based on modeling sequences of context conditions, the knowledge of the long and short-term context can contribute to producing an accurate recommendation model with a relatively small number of contextual conditions. In order to investigate the effect of sequence length on the recommendation accuracy, we generated sequences of different lengths and applied SLCM with the CARS dataset. As can be seen from the results presented in figure \ref{fig:RMSE_SLCM}, the prediction results of shorter sequences (i.e., two, three, and four) produce better results in terms of the RMSE and MAE. We can also see that the best results of 0.336 for the RMSE measure were achieved in sequence lengths of two and three, when taking into account the sequences composed of the user's last one or two contextual interactions. As for the MAE measure, extracting latent context from sequence lengths of two and four obtained the best results (0.243). However, the RMSE and MAE results in the setting of longer sequences (i.e., five and six) were higher, as the RMSE and MAE increased to 0.344 and 0.275 respectively (compared to the setting with shorter sequence length). 

\section{Conclusions}
\label{sec:conclusion}
We presented deep context-aware recommendation models that utilize explicit and latent context representations and learn nonlinear interaction function between users, items, and contexts. 
We conducted several experiments on two context-aware datasets to compare our approaches to state of the art recommendation models with respect to the hit@k, RMSE, and MAE measures. We showed that our models significantly outperform popular baselines used for rating prediction task. 
Overall, the sequential latent context model produced better results than the other models. In terms of ranking quality, SLCM improved baseline models up to 16.3\% in the hit@k measure. With regard to prediction accuracy, SLCM improved baseline models up to 9.47\% for the RMSE measure and 8.98\% for the MAE measure.     
During the experiments, we performed sensitivity analysis to determine the optimal sequence length and observed that sequences composed of a limited number of context vectors produce satisfactory results. 
In future work, we intend to extend this study by including groups of users in order to solve the new user problem. We also plan to integrate rich item representations that are derived from the item's content (e.g., thumbnail, text) and examine other network structures e.g CNN which contains less parameters.


\bibliographystyle{main}
\bibliography{main}

\end{document}